\documentclass{article}
\usepackage{PRIMEarxiv}

\usepackage[utf8]{inputenc} 
\usepackage[T1]{fontenc}    
\usepackage{hyperref}       
\usepackage{url}            
\usepackage{booktabs}       
\usepackage{amsfonts}       
\usepackage{nicefrac}       
\iftutex
  \usepackage{mathtools}
  \usepackage{unicode-math}
  \defaultfontfeatures{Scale=MatchLowercase}
  \setmathfontface{\mathui}{CMU Serif Upright Italic} 
\else
  \usepackage{lmodern}
  \usepackage{mathtools}
  \usepackage{amsfonts}
  \DeclareMathAlphabet{\mathui}{T1}{cmr}{m}{ui}
\fi

\usepackage{microtype}      
\usepackage{lipsum}
\usepackage{graphicx}
\graphicspath{{media/}}     

\usepackage{amsmath}
\usepackage{amssymb}
\usepackage[dvipsnames, table]{xcolor}
\definecolor{forestgreen(web)}{rgb}{0.13, 0.55, 0.13}
\definecolor{antiquefuchsia}{rgb}{0.57, 0.36, 0.51}
\usepackage[]{mdframed}
\usepackage{multirow}

\usepackage{amsthm}
\usepackage{makecell}

\usepackage{tabularx,multicol,booktabs,multirow,makecell}[width=\textwidth]

\theoremstyle{definition}

\title{An explainable approach to detect case law on housing and eviction issues within the HUDOC database
}

\author{M. Mohammadi  \\
        University of Groningen\\
        The Netherlands\\
        \texttt{mohammadimathstar@gmail.com}\\
        \And
        M. Wieling \\
        University of Groningen\\
        The Netherlands\\
        \texttt{m.b.wieling@rug.nl}\\
        \And
        M. Vols \\
        University of Groningen\\
        The Netherlands\\
        \texttt{m.vols@rug.nl}\\
}

\begin{document}
\maketitle

\begin{abstract}
Case law is instrumental in shaping our understanding of human rights, including the right to adequate housing. The HUDOC database provides access to the textual content of case law from the European Court of Human Rights (ECtHR), along with some metadata. While this metadata includes valuable information, such as the application number and the articles addressed in a case, it often lacks detailed substantive insights, such as the specific issues a case covers. This underscores the need for detailed analysis to extract such information. However, given the size of the database - containing over 40,000 cases - an automated solution is essential.

In this study, we focus on the right to adequate housing and aim to build models to detect cases related to housing and eviction issues. Our experiments show that the resulting models not only provide performance comparable to more sophisticated approaches but are also interpretable, offering explanations for their decisions by highlighting the most influential words. The application of these models led to the identification of new cases that were initially overlooked during data collection. This suggests that NLP approaches can be effectively applied to categorise case law based on the specific issues they address.


\end{abstract}

\keywords{Right to housing \and Eviction \and Prototype-based learning \and Document classification \and Explainable AI}

\section{Introduction}


Interpreting human rights in light of current conditions and developments in international law is crucial, requiring legal scholars to continuously follow these changes \cite{letsas2013echr}. In this context, courts' decisions play a fundamental role in legal research. Legal scholars often rely on these decisions to track the evolution of human rights interpretation, which adapts in response to societal changes and international consensus. Consequently, scholars frequently need to retrieve case law relevant to specific issues, such as housing problems. 

The European Convention on Human Rights (ECHR) and the European Court of Human Rights (ECtHR) play an important role in the interpretation and protection of human rights across Europe. While the ECHR provides a framework for protecting fundamental rights, the ECtHR ensures these rights are enforced in real life by interpreting and applying the convention in specific cases. Thus, the HUDOC database, comprising the decisions of the ECtHR, is an essential resource for legal scholars studying human rights in Europe. It provides both textual content of cases and some metadata. While metadata contains useful information, such as the application number, articles, and citations, it does not have information such as what substantive issue a case law covers which makes it difficult for a scholar who is interested in a specific topic. Due to the size of the database, it is demanding to find efficient ways to extract such information. 

In this paper we focus on one of these specific topics, namely housing. In the context of housing issues, previous research has shown that Article 8 and Article 1 of Protocol 1 are particularly relevant, with the Court’s decisions shaping the legal landscape of housing rights\cite{bruijn2024navigating}.  Given that most housing-related cases are found under Article 8 and Article 1 of Protocol 1, we have collected several hundred relevant cases on housing and eviction (i.e.~a subset of the housing cases). While this dataset serves as a valuable resource for empirical studies, many relevant cases are very likely to be absent. Examining all cases in HUDOC to obtain a comprehensive dataset is infeasible due to the large number of cases, necessitating a more practical approach to efficiently retrieve relevant cases.

With advancements in computational technology and the existence of large legal databases, Natural Language Processing (NLP) techniques are increasingly employed to analyze large databases. Topic modelling models have been used across various jurisdictions to explore various legal databases \cite{carter2016reading,remmits2017finding,luz2020topic,aguiar2022using,novotna2020topic,silveira2021topic,razon2022topic,salaun2022tenants,mohammadi2024combining}. Additionally, several works have developed NLP models to classify legal documents. For instance, they have created automated systems to identify, categorize, and forecast court decisions \cite{aletras2016predicting,medvedeva2020using,medvedeva2021automatic,medvedeva2021automatically,medvedeva2023rethinking}, 
or have been used to forecast future citations to case law \cite{mones2021emergence,schepers2023predicting} and to summarize legal cases using large language models (LLM) \cite{pandya2019automatic,kanapala2019text,galgani2015summarization,kumar2012legal,kim2013summarization}.

In this paper, our objective is to develop automated models capable of retrieving cases on housing and eviction issues within the HUDOC database. First, we utilized an Adaptive Chordal Distance-based Subspace Learning Vector Quantization (AChorDS-LVQ) model to build a classifier for legal document classification. We then demonstrate how the model provides insights into its decisions. Additionally, we compared our model with transformer-based language models (LM), highlighting the advantages and trade-offs of each approach. Finally, we applied our models to the entire corpus of ECtHR cases to detect new relevant cases. Specifically, we initially focused on cases addressing Articles 8 and Article 1 of Protocol 1. Later, we evaluated whether the trained model could generalize its knowledge to classify cases beyond these articles. 

This contribution is organized as follows: Section \ref{sec:data} explains how we collect the required data for training models. Section \ref{sec:method} provides a brief introduction to the AChorDS-LVQ. Section \ref{sec:experiment} covers the experiments where we developed models for case law classification and then used them to detect more relevant cases. We finally end with a conclusion in Section~\ref{sec:conc}.  

\section{Data}\label{sec:data}
For this paper, we collected all cases (judgments and decisions) within the HUDOC database published up to and including January 2023 for which (English) texts are available. This yields 41,621 cases. 
As the most relevant articles within the ECHR concerning housing-related issues are Article 8 (right to respect for private, family life, home and correspondence) and Article 1 of Protocol No.~1 (right to protection of property), we focused on these (13,343) cases for the creation of a dataset for building an automated system to detect housing-related cases within the case law of the ECtHR. 

Following the methodology introduced in \cite{mohammadi2024combining}, we employed a combination of textual analysis and citation patterns to collect a set of cases potentially addressing housing issues.
From this procedure, we collected 1,108 potentially housing-related cases. 
Then, each decision/judgement was annotated manually to determine its relevance to housing issues and whether it addressed eviction specifically. The annotaters looked in the facts of the case and the complaint of the applicant. If the facts or complaint predominantly concern someone's house, home, residence or accommodation the case was characterized as housing-related. The same applies if the case dealt with issues regarding landlords and tenants. The annotators also coded whether the case is eviction-related or not. Eviction was defined as the loss of one's home, and a tenure-neutral broad definition was applied. As a result, tenants, homeowners, squatters and illegal occupiers could all be subject to an eviction. The annotaters were trained and used a code book. The inter-annotator agreement for these variables was measured using Fleiss’ kappa score. The kappa values varied between 0.81 and 0.89, which can be considered to be (almost) perfect (0.81–1.00) agreement levels.  

As expected, the majority (78\%) of the 1,108 cases in our manually annotated dataset was housing-related, resulting in an imbalanced dataset. This can lead to the existence of bias in the model toward predicting cases as a housing-related case. 
To reduce this effect, we randomly selected 507 additional cases linked to Article 8 ECHR or Article 1 of Protocol No.~1 ECHR which were not found to be related to housing-related cases using our textual analysis method \cite{mohammadi2024combining}. A summary of the resulting dataset is presented in Table \ref{tab:dataset}. While the majority of the new cases are indeed unrelated to housing issues, there are still quite a few cases on housing that were not identified previously (12\%). This shows that unsupervised methods, such as those in \cite{mohammadi2024combining}, may miss relevant cases. 
Therefore, it is necessary to assess whether new (supervised) models can be developed that identify the characteristics of housing cases through a machine learning approach.

\begin{table}[]
    \centering
    \begin{tabular}{c|c | c|c}
        \hline
        Method &Following & Random & \\
        (for collection) &\cite{mohammadi2024combining} & &Total\\
        \hline
        Housing & 865 & 63 & 928\\
        Not-housing & 243 & 444 & 687\\
        \hline 
        Eviction & 678 & 26 & 704\\
        Not-eviction & 430& 481& 911 \\
        \hline
        Total &1108& 507 \\
    \end{tabular}
    \caption{Summary statistics of the collected dataset. Note that the eviction cases consists of a subset of the housing cases.}
    \label{tab:dataset}
\end{table}

We therefore aim to utilize this manually annotated dataset of case law to build classifiers predicting the case law labels, specifically trying to identify those related to housing and eviction. This dataset will be used for training (80\%) and then testing (20\%) our automated systems for detecting relevant cases to housing and eviction within the HUDOC database. 

\section{Method} \label{sec:method}
In this section, we present an overview of the Adaptive Chordal Distance-based Subspace Learning Vector Quantization (AChorDS-LVQ) method, which we use to classify case law. To achieve this, we begin with a brief introduction to the Generalized Learning Vector Quantization (GLVQ) framework, the foundation for the AChorDS-LVQ method. We then give a summary of AChorDS-LVQ which combines the power of word embedding models with the explainability of GLVQ, offering a robust solution for applications that require both accuracy and interpretability.


\subsection{Generalized Learning Vector Quantization} \label{subsec:GLVQ}

\begin{figure}
    \centering
    \includegraphics[scale=0.7]{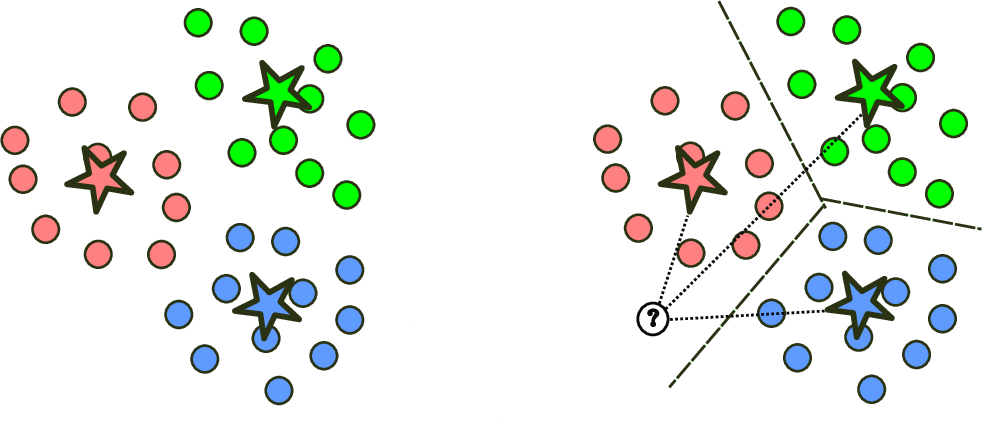}
    \caption{A conceptual visualization of LVQ classifiers. Left: training examples (circles) and learned prototypes (stars) with colours as labels. Right: illustration of the Nearest Prototype Classifier (NPC) strategy with an uncoloured circle as an unseen example. Note that the dashed lines represent borders between classes and dotted lines denote the difference between an unseen example and prototypes. 
    }
    \label{fig:lvq-visualization}
\end{figure}

Generalized Learning Vector Quantization (GLVQ) \cite{sato1995generalized} represents a family of classifiers known for their intrinsic ability to explain their decisions. It utilizes a set of labelled prototypes, which are representative examples summarizing the characteristics of each class. 
Because of its transparency, it has been applied in different fields where it is necessary to have a transparent model, ranging from healthcare \cite{ghosh2020visualisation, ghosh2022interpretable, hankel2017sequence} and education \cite{widiantara2023application}, to astronomy \cite{mohammadi2022detection}.


Let $\{ (\vec{x}_i, y_i) \}_{i=1}^N$ be the training set where $\vec{x}_i \in \mathrm{R}^D$ is a training example and $y_i$ is its corresponding label. Given the training set, a GLVQ classifier models classes through a set of labelled prototypes, i.e.:
\[
\{ (\vec{w}_i, c(\vec{w}_i) \}_{i=1}^p
\]
where $\vec{w}_i \in \mathrm{R}^D$ denotes a prototype vector and $c(\vec{w}_i)$ is its corresponding label. As can be seen in the left visualization presented in Fig.\ \ref{fig:lvq-visualization}, prototypes capture major characteristics of each class and their distribution simulates the distribution of classes within the data.
During training, the classifier learns the best places for prototypes by optimizing the following cost function
(more details in \cite{sato1995generalized}).
\begin{equation}
    E = \sum_{i=1}^N E_i = \sum_{i=1}^N \text{sgd} \Big( \frac{d(\vec{x}_i, \vec{w}^+) - d(\vec{x}_i, \vec{w}^-)}{d(\vec{x}_i, \vec{w}^+) + d(\vec{x}_i, \vec{w}^-)} \Big) \enspace,
    \label{eq:cost-fun}
\end{equation}
where sgd is the sigmoid function, and $d(., .)$ represents a distance measuring the difference between an example and a prototype (dotted lines in Fig.~\ref{fig:lvq-visualization},  right). Note that one can interpret $E_i$ as the probability of misclassifying the i-th example. Consequently, the model aims to reduce the overall probability of misclassifications.

Once the model learns the optimal prototypes, it follows the Nearest Prototype Classifier (NPC) strategy to perform predictions for unseen examples (see Fig.~\ref{fig:lvq-visualization}, right). This means that for a new example $\vec{x}$, we first find the closest prototype to it. Once we find the closest prototype, we assign its label (red colour in the example shown in Figure~\ref{fig:lvq-visualization}, right) to $\vec{x}$. In the context of case law, it will classify a case as housing related, only if it's closest to the prototype representing housing-related cases (and not another prototype). 

Since prototypes are positioned within the data space ($\mathbb{R}^D$), so the complete set of cases, they are highly explainable and one can easily identify the major characteristics within classes (e.g., through shared concepts). Due to GLVQ's explanatory power, different variants of it have been developed to enhance its flexibility and explainability. These variants, including AChorDS-LVQ, differ by proposing different distance measures $d$. 

\subsection{Adaptive Chordal Distance-based Subspace Learning Vector Quantization} \label{subsec:GRLGQ}
The first step towards the classification of textual data such as case law of the ECtHR is to convert the textual data to a numerical representation. In other words, we need to transform words into numbers such that these numbers capture the meaning of the words. In simpler terms, if two words have similar meanings, their numerical representations (i.e.~two series of numbers contained in two vectors) also need to be close to each other.
With the advancement in word embedding techniques, several approaches have been developed to capture the semantic meaning of words in numerical vectors. This has started with Word2Vec \cite{mikolov2013efficient} and GloVe \cite{pennington2014glove} which use the co-occurrence of words to determine similarity (e.g., `cat' will more often co-occur with `dog' than with `judge', and therefore the vector representations of `cat' and `dog' will be more similar than that of `cat' and `judge'). 
These approaches therefore are an efficient and effective way to represent words and text through numbers. 

Let $\text{doc} = (\text{word}_1, \text{word}_2, \cdots, \text{word}_n)$ be a text with $n$ words. A word embedding model assigns a vector representation to each word, resulting in the text being represented as a set of vectors:
\[
(\text{word}_1, \text{word}_2, \cdots, \text{word}_n) \xrightarrow[]{\text{GloVe}} (v_1, v_2, \cdots, v_n)
\] 
where $v_i$ is the numerical representation for the i-th word $\text{word}_i$, called an embedding vector. 

The conventional machine learning approaches, such as Support Vector Machines (SVM) and GLVQ, accept one vector per example as input. To apply these methods to textual data, it is therefore necessary to summarize an entire text into a single vector instead of a set of vectors. A common practice to prepare data is to use the mean vector, i.e. $\frac{1}{n} \sum_{i=1}^n v_i$. However, this summarization can result in losing valuable information needed to differentiate documents effectively. 

To address this issue, AChorDS-LVQ \cite{mohammadib2024nips} instead uses a set of vectors to represent a text. This approach aligns better with the nature of texts, which consist of many words rather than a single word, thus preserving more information and improving classification accuracy.
Since the number of words $n$ in documents can vary between documents, AChorDS-LVQ employs a dimension reduction technique, Singular Value Decomposition (SVD), to represent each document with a fixed number of vectors $d$:
\[
(\text{word}_1, \text{word}_2, \cdots, \text{word}_n) \xrightarrow[]{\text{GloVe}} (v_1, v_2, \cdots, v_n) \xrightarrow[]{\text{SVD}}  (u_1, u_2, \cdots, u_d) 
\] 

Being a member of the GLVQ family, AChorDS-LVQ defines a set of labeled prototypes $\{(W_i, y_i)\}_{i=1}^p$. Unlike GLVQ, a prototype $W_i$ consists of $d$ vectors, instead of a single vector. The flexibility introduced by using several vectors for prototypes and documents results in a document-classification-model achieving comparable results to Large Language Models (LLM) models. However, it requires less computational power \cite{mohammadib2024nips}, and more importantly, AChorDS-LVQ provides insights about its decisions by quantifying the effect each word has on its classification. As such, the model's logic is well-aligned with human expectations, as one can inspect which words have the highest impact on the model's decisions.


\section{Experiment}\label{sec:experiment}

In the following, we aim to develop classifiers to identify cases dealing with housing and eviction issues from Article 8 ECHR and Article 1 of the First Protocol ECHR. First, we create and evaluate models that classify case law based on whether they concern a housing issue or not on the basis of our constructed dataset. 
Then, we used the obtained models to classify new cases not in our constructed dataset to identify potential new housing-related cases. Specifically, we first apply our models to unseen cases addressing Article 8 and Article 1 of Protocol No.~1. This is followed by using our models to identify housing-related cases from the set of cases not linked to these articles. This second approach aims to evaluate how much the knowledge learned by models from Article 8 and Article 1 of Protocol No.~1 is transferable to identifying housing-related case law from the other articles in the ECHR.

\subsection{Experimental Setup}
Our goal is to develop binary classifiers to label case law documents as either not addressing a housing-related issue (0), or addressing a housing-related issue (1). Similarly, we create binary classifiers labeling case law documents as eviction-related or not.
In order to generate the training and test sets, we randomly select 80\% of the cases in our manually annotated dataset as training examples, and we use the rest of the dataset to evaluate our approach. This results in 1,292 cases used for training (728 on housing and 561 on evictions) and 323 cases for testing (200 on housing and 143 on evictions).

Given the limited amount of annotated data available (fewer than 2000 cases), we use two strategies for building our classifiers:

\begin{itemize}
    \item Learning from scratch: this approach involves training models from the ground up using our annotated dataset. However, due to the small size dataset, only models with lower complexity are feasible, such as SVM, GRLGQ\cite{mohammadi2024arxiv} and AChorDS-LVQ\cite{mohammadib2024nips}.
    \item Fine-tunning Large Language Models (LLM): Through technological advances, we have access to more complex models with millions (or billions) of parameters.
    These models are trained on large corpora of texts, seemingly obtaining somewhat of an understanding of the structure of human language. These models can subsequently be fine-tuned with smaller datasets for specific tasks, such as in this case classifying legal documents.  
\end{itemize}
In this study, we trained SVM, GRLGQ\cite{mohammadi2024arxiv} and AChorDS-LVQ\cite{mohammadib2024nips} models from scratch. Additionally, we fine-tuned three large pre-trained models: a) Bidirectional Encoder Representations from Transformers (BERT) pre-trained on general texts, b) Legal-BERT pre-trained on legal data, and c) Longformer pre-trained on general texts but designed to accept longer texts. 

To train our models, we first need to prepare the documents appropriately. For simpler models, we must convert the text into numerical representations. A common method for this is to use word embedding models, such as Word2Vec and GloVe.
Following the approach outlined in \cite{mohammadib2024nips}, we use the GloVe model to represent the text. The preprocessing steps are as follows:
\begin{enumerate}
    \item Stop words removal: We begin by removing all stop words (i.e.~uninformative function words), such as `and' and `is', to reduce the influence of common words that add little value to the document's meaning.
    \item Word embedding: Using the GloVe model, we generate a vector representation for each word in the document.
    \item Case law representation:
    \begin{itemize}
        \item SVM: We represent each document by calculating the mean vector of all the word vectors generated by GloVe.
        \item GRLGQ and AChorDS-LVQ: For these models, we apply Singular Value Decomposition (SVD) to the word vectors to generate the appropriate document representation with multiple vectors.
    \end{itemize}
\end{enumerate}
This process generates the required input for SVM, GRLGQ and AChorDS-LVQ.

Due to the computational complexity, transformer-based language models, such as BERT, have a limit on the length of text they can process, known as the context window. For instance, BERT and LegalBERT are restricted to a maximum of 512 tokens. There are several approaches to address this limitation. A common approach to tackle this challenge is to segment the long text into smaller pieces and then perform prediction for each piece separately. Following that approach, we split documents here into non-overlapping chunks of 300 words. During training, we assign to each chunk the document's label and then use it to train BERT and LegalBERT. At test time, the model predicts labels for each chunk individually, and then the document's predicted label is the one that appears most frequently among its corresponding chunks. 
When we report the results, we add the subscript `long' to the model names, to distinguish this adjustment.

Another approach to address the limited context window is to find a way to reduce the computational cost. For this, the Longformer model has been introduced \cite{beltagy2020longformer}. This model can handle documents with 4096 tokens. As more than 50\% of documents in our dataset are shorter than 2300 words, we directly use Longformer on our data without any splitting.

\subsection{Model performance}

\begin{table}[]
    \centering
    \begin{tabular}{c  | c | c c }
        \hline
        Method &  \# of para.&Housing& Eviction\\
        \hline
        SVM & 256k, 260k&87.62& 82.97\\
        GRLGQ & 30k&93.11 ($\pm 0.11$)&90.92 ($\pm0.15$)\\
        AChorDS-LVQ& 30k& \textbf{93.60} ($\pm 0.15$)& 91.53 ($\pm 0.39$)\\ 
        \hline
        BERT & 110M & 91.33 & 87.62\\
        LegalBERT & 110M & 92.57 & 88.85\\
        BERT$_\textrm{long}$ & 110M & 93.17 & 91.64\\
        LegalBERT$_\textrm{long}$ & 110M & 93.48 & \textbf{92.24}\\
        \hline
        Longformer &148M& 93.19 &91.64\\
        \hline
    \end{tabular}
    \label{tab:accuracies}
    \caption{Performance of different methods on both binary classification tasks.}
\end{table}

Table~\ref{tab:accuracies} shows the performance of the different approaches in the classification of case law. For the classifier that detects housing-related cases, we see that the AChorDS-LVQ and GRLGQ provide comparable results to the transformer-based models, with the more computationally efficient AChorDS-LVQ yielding the best performance. 
This suggests that AChorDS-LVQ and GRLVQ offer a cost-effective solution for creating models to detect relevant cases, making them suitable for applications using large legal databases.

For the eviction classifier, all models achieve a lower performance compared to the housing classifiers. This difference is due to the smaller number of eviction-related examples in the training set. In this context, LegalBERT$_\textrm{long}$ outperforms the others. One possible reason for this is that LegalBERT, which was pre-trained on legal data, already has a general understanding of legal texts, enabling it to achieve a higher accuracy even with small datasets. Despite providing slightly lower accuracy, AChorDS-LVQ still delivers acceptable results, offering a good balance between accuracy and computational cost. As it also allows for explainable results, we have opted to use AChorDS-LVQ in our analysis of the HUDOC database.

As previously mentioned, AChorDS-LVQ is a transparent model, which means that it specifies the impact of each word from the case law document on its final decision. To illustrate this, we consider one case from each class and demonstrate which words have a high impact on the model's decision.

\begin{figure}
    \centering
    \includegraphics[scale=0.8]{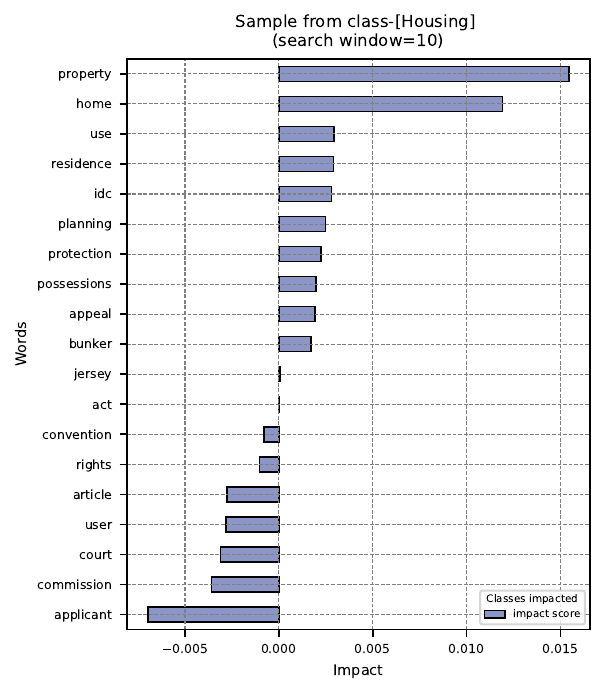}
    \includegraphics[scale=0.8]{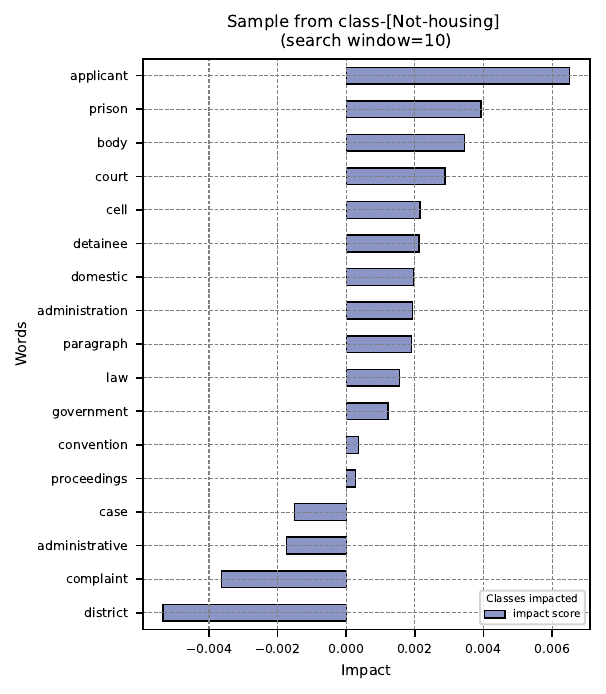}
    \caption{Top words with their impact on the classifier's decision. Left: a housing-related case - ECHR 11 March 1985, no. 11185/84 (\textit{Muriel HERRICK v. UK}). Right: a non-housing related case - ECHR 13 November 2014, no. 1088/10 (\textit{MERZAĻIJEVS v. LATVIA}).}
    \label{fig:housing-explainability}
\end{figure}

Specifically, we applied the model to the decision with application number 11185/84. The model correctly predicts that the case involves a housing-related issue, and it also assigns a numerical value to each word, indicating its influence on our model's decision. The left graph shown in Fig.\ \ref{fig:housing-explainability} displays several words with the highest impact. The blue bar represents the extent to which a word supports the correct class (i.e.~`housing'). The larger (i.e.~more towards the right) the blue bar, the more important the word is for the correct class `housing'. Conversely, the smaller (i.e.~more towards the left) the blue bar, the more important the word is for the incorrect class `non-housing'. It can be seen that words such as property, home, residence, and bunker are important for housing-related cases, aligning well with our intuition. Additionally, we observe that while housing-related terms correctly support the claim that the case covers a housing issue, procedural terms such as `convention', `article', and `court' do not.


Similarly, we utilized the model to analyze the ECtHR's decision with application number 1088/10. The model correctly classifies this case as non-housing-related. The right graph shown in Fig.\ \ref{fig:housing-explainability} highlights the most important words influencing this prediction. For this particular case, words such as `prison', `cell' and `detainee' are highlighted by the model. These terms suggest that this case is related to prison conditions, thereby reinforcing the model's classification. This detailed analysis confirms the model's capability to accurately interpret and classify case law based on the specific terminology used, thereby also showcasing its reliability and transparency.

\begin{figure}
    \centering
    \includegraphics[scale=0.75]{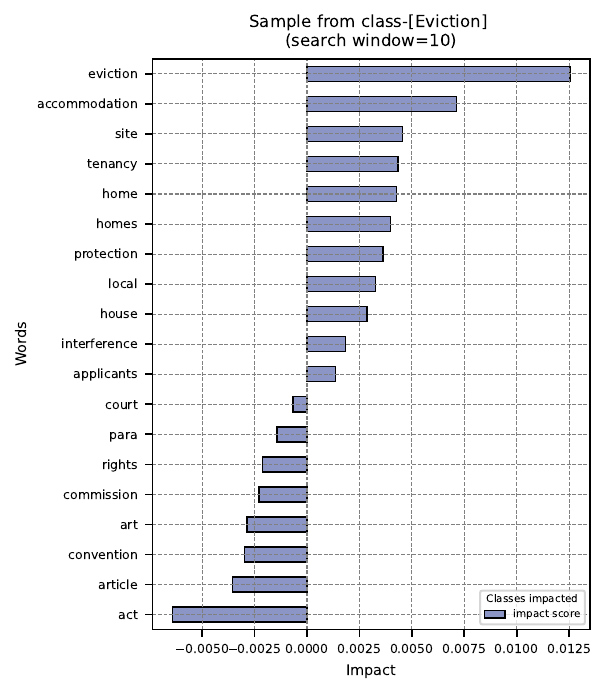}
    \includegraphics[scale=0.75]{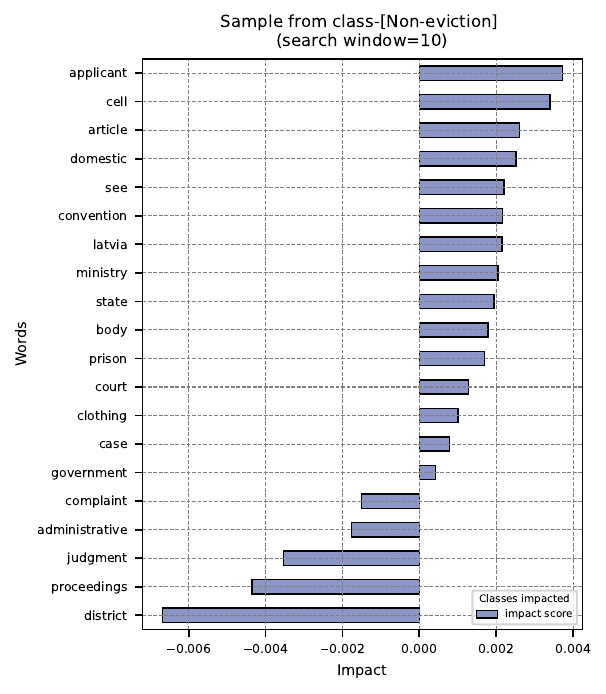}
    \caption{Top words with their impact on the classifier's decision. Left: for an eviction-related case - ECHR 12 December 1990, no. 14751/89 (\textit{P. v. UK}, date 12.12.1990). Right: for a non-eviction-related case - ECHR 13 November 2014, no. 1088/10 (\textit{MERZAĻIJEVS v. LATVIA}).}
    \label{fig:eviction-explainability}
\end{figure}


To evaluate the model developed for detecting eviction-related cases, we applied it to two correctly classified examples: one related to eviction and the other unrelated. Note that the second example is the same as the one used for the evaluation of the housing model to illustrate the different important words. Fig.\ \ref{fig:eviction-explainability} highlights the most impactful words influencing the model's decisions.

For the eviction-related case (left), as expected, words such as `eviction', `accommodation', `site', `tenancy', and `home(s)' play a major role in the model's decision to classify the judgement as eviction-related. This shows that the eviction prototype ($W^{\text{evict}}$) accurately encodes the characteristics of eviction issues by capturing the most representative words for this class. 
Conversely, for the non-eviction case (right), the model highlights words related to `prison conditions', which are not relevant to eviction, as the most influential. This further demonstrates the model's ability to distinguish between eviction-related and unrelated cases effectively.

\subsection{Detection of new housing-related cases within Article 8 and Article 1 of Protocol No.~1}
In the HUDOC database, there are 13,343 cases addressing Article 8 or Article 1 of Protocol 1 ECHR. As indicated, we have manually coded 1,615 cases of these cases (12\%). To identify other housing-related cases, we applied the same model, used in the previous section, to all these cases. This resulted in the detection of 1,738 cases which are potentially related to housing issues. 
Given the large number of newly detected cases and the time-consuming process of manual annotation, it may be useful to prioritize the cases with the highest likelihood of being housing-related. We therefore developed a way to convert the model’s outputs into probability scores. These scores represent the likelihood that a given case is related to housing, allowing us to rank the cases accordingly. 
As a result, we can more effectively focus on the most relevant cases. 

Following the definition of the cost function in Equation \ref{eq:cost-fun}, we use the calculated distances to prototypes to define the probability of a case being related to housing as follows:
\begin{equation}
    P(\text{case is on housing}) = \text{sgd} \left(\frac{d(\vec{x_i}, \vec{w}^{\text{NH}}) - d(\vec{x_i}, \vec{w}^{\text{H}})}{d(\vec{x_i}, \vec{w}^{\text{NH}}) + d(\vec{x_i}, \vec{w}^{\text{H}})}\right)
    \label{eq:score_housing}
\end{equation}
where $d(\vec{x_i}, \vec{w}^{\text{NH}})$ and $d(\vec{x_i}, \vec{w}^{\text{H}})$ represent the distance between the document $\vec{x}_i$ and the prototypes of non-housing (NH) and housing (H), respectively.

\begin{figure}
    \centering
    \includegraphics[scale=0.46]{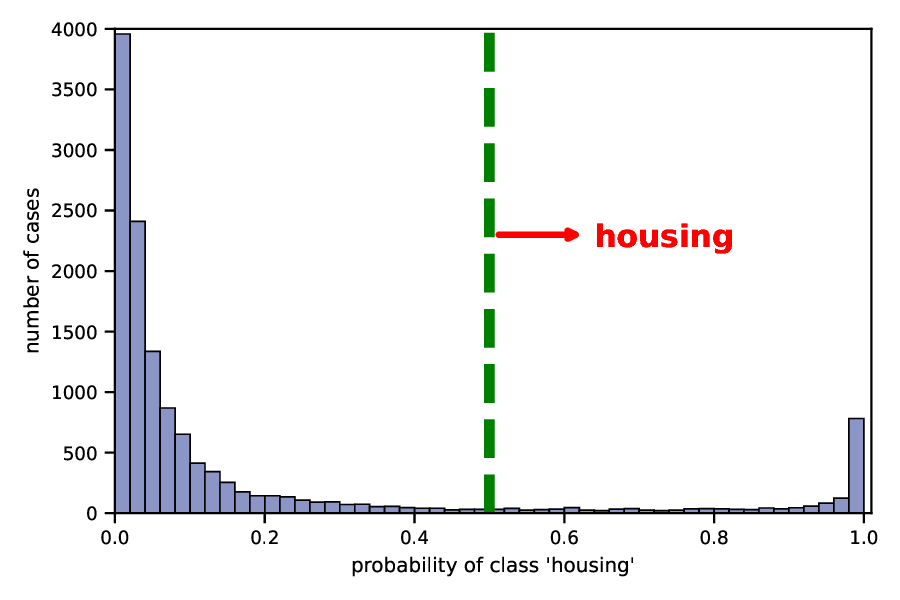}
    \includegraphics[scale=0.45]{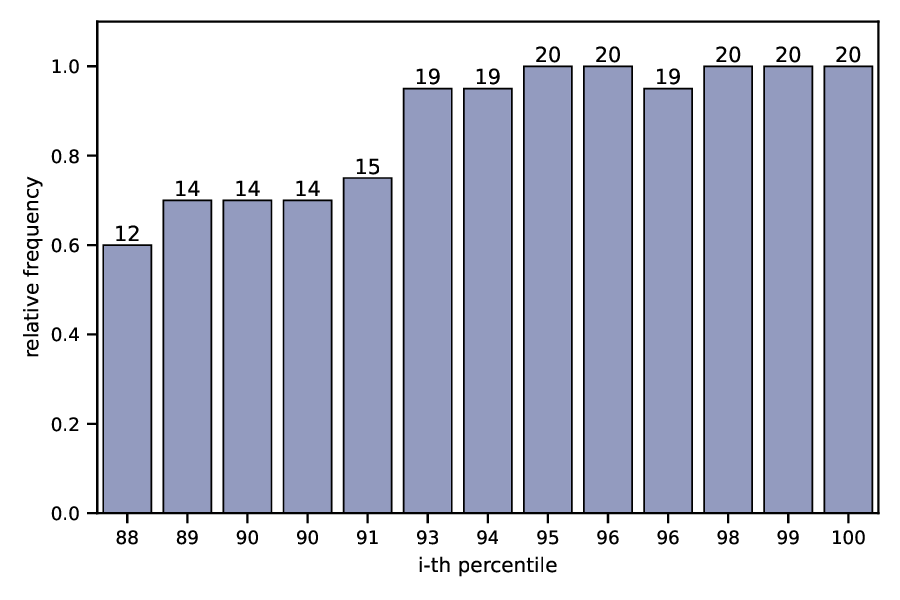}
    \caption{Identified cases related to housing for Article 8 and Article 1 of Protocol 1. Left: the distribution of probability scores indicating the likelihood of being related to housing. Right: the fraction of randomly selected top-13-percentile cases which were manually annotated to be housing-related.}
    \label{fig:scores_housing_8p1}
\end{figure}

The left graph shown in Fig.\ \ref{fig:scores_housing_8p1} shows the distribution of probability scores for cases being housing-related. As expected, most cases receive low scores, indicating that they are most likely not related to housing. Using these scores, we can set a threshold to identify cases with a high likelihood of being housing-related. By default, the model predicts a case to be about housing if its score is above 0.5.

To identify the appropriate threshold, we have calculated the percentile associated with each score. In this way, we found that the 88th percentile was associated with a score of 0.502. We then randomly selected and annotated 20 cases from each of the top 13 percentiles. The right graph of Fig.\ \ref{fig:scores_housing_8p1} shows the results of the manual check. Specifically, these results illustrate a direct relationship between the scores and the likelihood of being a housing-related case. Setting a high score threshold will thus result in most of the selected cases being housing-related, whereas a lower threshold may include more non-related cases. For example, setting the score threshold to 0.91 would result in cases which are more than 95\% likely to be housing-related. Of course, more cases will be selected when a lower threshold is used, so it is useful to adjust the threshold based on the availability of human annotators.

In summary, this analysis highlights the model's effectiveness in uncovering relevant cases that might have otherwise been overlooked. Consequently, this method may help in providing a more accurate and thorough understanding of the application and interpretation of Article 8 and Article 1 of Protocol 1 in housing-related matters as it is able to more completely identify housing-related cases.


\subsection{Detection of new cases from other articles in the ECHR}

While Article 8 and Article 1 of Protocol No.~1 ECHR are typically the primary articles addressing housing-related issues, other articles may also deal with housing and eviction-related issues. Thus, to improve the comprehensiveness of our housing-related dataset, we applied our model also to the remaining HUDOC database, which comprises 28,278 cases, aiming to detect potentially housing-related cases from other articles. Among these, the model predicted 567 new cases (with probability scores higher than 0.5) that are likely related to housing.

\begin{figure}
    \centering
    \includegraphics[scale=0.46]{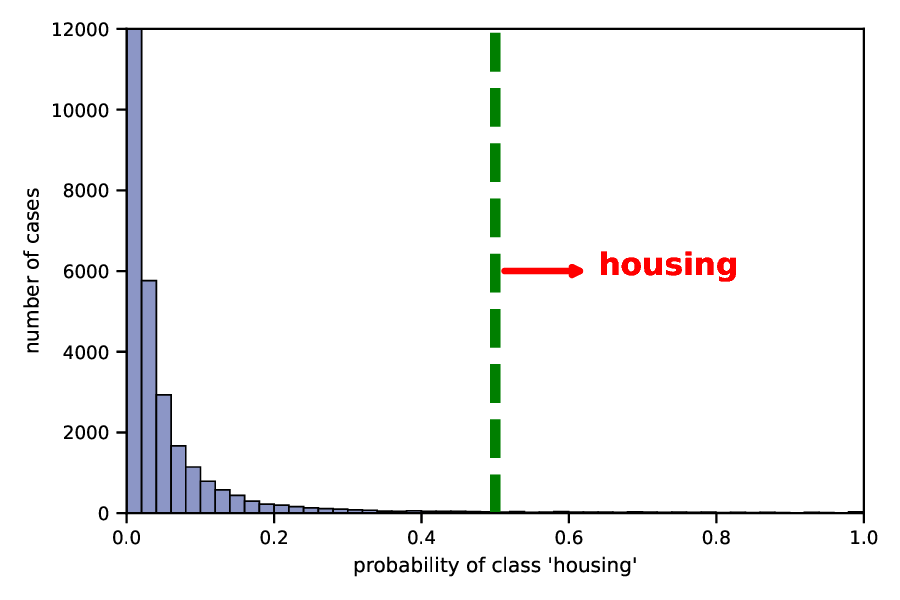}
    \includegraphics[scale=0.45]{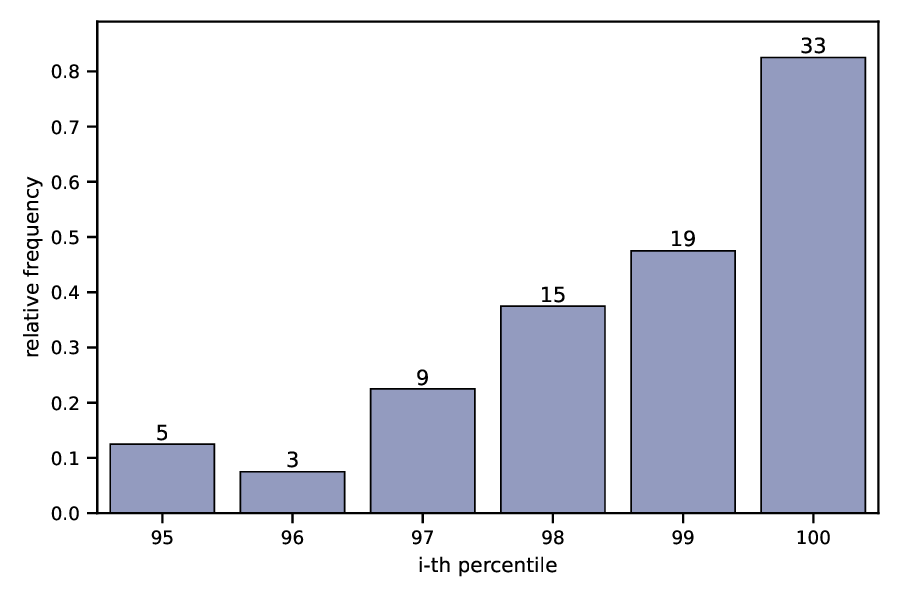}
    \caption{Identified cases related to housing for cases not related to Article 8 and Article 1 of Protocol 1. Left: the distribution of probability scores indicating the likelihood of being related to housing. Right: the fraction of randomly selected top-six-percentile cases which were manually annotated to be housing-related.}
    \label{fig:scores_housing_others}
\end{figure}

Similar to the previous experiment, we computed the probability scores representing the likelihood of cases being related to housing. The left graph shown in Fig.\ \ref{fig:scores_housing_others} illustrates the distribution of these scores. The majority of cases receive low scores, with only the top two percentiles receiving scores above 0.5.

To assess the model's performance across different score ranges, we randomly selected and annotated 40 cases (per percentile) from the top six percentile (i.e.~scores greater than 0.21). The right graph shown in Fig.\ \ref{fig:scores_housing_others} presents the annotated results. As expected, the model's performance is lower for these cases compared to those under Article 8 and Article 1 of Protocol 1, reflecting the specific training focus. Nevertheless, this model remains valuable by detecting relevant cases from other articles that might otherwise be overlooked, particularly if a higher threshold is used (i.e.~using a threshold of 0.71 would result in an expected 82.5\% of cases being related to housing). 

Additionally, we observe that the probability of a case being housing-related generally also appears to decrease as scores decrease (i.e.~a score of 0.52 has only a 47.5\% chance of being housing-related, whereas this is only 7.5\% for a score of 0.25). We also observe that the percentage of correctly classified relevant cases experiences a sharp drop from 82.5\% (for cases with scores above 0.71) to 47.5\% (for cases with scores between 0.52 and 0.71). This shows the importance of setting a higher threshold to focus on cases the model is more confident about (i.e.~above 0.71), although cases with lower scores can still be reviewed when human annotators are available (as the probability of identifying housing-related cases, even for the percentiles associated with lower scores is greater than 0).

This experiment again shows the model's utility in identifying and prioritizing cases based on their likelihood of being relevant, enabling legal scholars to identify housing-related decisions across various legal articles and thereby gaining deeper insights into judicial decisions when analyzing this larger set of cases.


\section{Conclusion}
\label{sec:conc}
As international human rights conventions are considered `living instruments', legal scholars must continuously track the developments of conventions to enhance our understanding of human rights. In this context, empirical research on court decisions plays an important role in our understanding of the interpretation and application of these conventions.
However, this task is challenging given the large legal databases, and it is therefore necessary to be able to automatically identify relevant case law on specific issues. In this study, we developed and evaluated models specifically designed to retrieve housing and eviction-related cases from the HUDOC database. Our primary objective was to improve the retrieval of relevant cases, thereby helping legal scholars in their analysis of housing rights within the European Convention on Human Rights (ECHR).

To achieve this, we first created a dataset on housing issues using an unsupervised process introduced in \cite{mohammadi2024combining}, which was manually corrected and extended. We then created various machine learning models to classify case law based on whether it covered the subject of interest (i.e.~housing). Through our experiments, we demonstrated that the AChorDS-LVQ model provides comparable performance to transformer-based models while offering more transparency and requiring a lower computational cost.

We then applied our models to the unannotated data within the HUDOC database, resulting in the detection of many new cases related to housing and eviction issues. Specifically, we identified 1.738 potentially new cases addressing Article 8 and Article 1 of Protocol No.~1 (our subsequent manual analysis of a subset of size 260 indicates that about 226 of these will actually be housing-related), and 567 cases from other articles are potentially housing-related (our manual analysis on 240 of these cases, reveals that 84 of these are expected to be actually housing-related). Importantly, our model provides scores that can be used to prioritize cases. This scoring system allows for a more efficient allocation of resources, ensuring that the most relevant cases are reviewed and annotated first. Our method is therefore quite effective in enriching datasets with new cases which are not easily identified through other (unsupervised) automatic means.

In conclusion, our work highlights the potential of the AChorDS-LVQ model in the legal domain, offering a practical and effective tool for legal case classification and retrieval. This approach enhances the comprehensiveness of empirical legal research on case law by helping the retrieval of more relevant cases. Future studies could further explore the applicability of AChorDS-LVQ to other legal classification tasks.

\bibliographystyle{unsrt}  
\bibliography{references}

\appendix
\clearpage
\renewcommand\thetable{\thesection.\arabic{table}}   
\setcounter{table}{1} 
\setcounter{page}{1}
\setcounter{section}{0}
\renewcommand{\thepage}{A\arabic{page}}
\renewcommand{\thesection}{A.\ \arabic{section}}  
\renewcommand{\thetable}{A.\ \arabic{table}}  
\renewcommand{\thefigure}{A.\ \arabic{figure}} 
\renewcommand{\theequation}{A.\ \arabic{equation}} 

\section{Appendix: Model specifications and computing resources}\label{A-sec:Compute resource}
\def\myColW{0.4\textwidth}
 \begin{table}[h!]
    \setlength\tabcolsep{1pt}
     \caption{The performance of different classifiers on document classification task.}\label{tab:doc_classification_perf}
     \begin{tabularx}{\linewidth}{@{\extracolsep{\fill}} 
            @{}>{\raggedright\footnotesize}p{0.16\textwidth}@{}*{2}{@{}>{\raggedright\footnotesize}p{\myColW}@{}}
            >{\footnotesize\arraybackslash}l@{} 
            }
\toprule
        \textbf{Model} & \textbf{Version and documentation link} &\textbf{Computing resource used} & 
 \\
\toprule
BERT&  \href{https://huggingface.co/google-bert/bert-base-uncased}{google-bert/bert-base-uncased }& T4GPU (via Google Collaboratory) & \\
LegalBERT&  \href{https://huggingface.co/nlpaueb/legal-bert-base-uncased}{nlpaueb/legal-bert-base-uncased}& T4GPU (via Google Collaboratory) & \\
Longformer& \href{ttps://huggingface.co/allenai/longformer-base-4096}{allenai/longformer-base-4096}& A100GPU (via Google Collaboratory) & \\
SVM & \href{https://scikit-learn.org/stable/modules/generated/sklearn.svm.SVC.html}{Sklearn SVM}& CPU & \\
GRLGQ & \href{https://github.com/mohammadimathstar/GRLGQ:}{Code repository of GRLGQ in Github}& CPU & \\
AChorDS-LVQ & Proposed method in this contribution. Code repository will be made public after acceptance of this work. & CPU & \\
\hline
word2vec& \href{https://radimrehurek.com/gensim/models/word2vec.html}{word2vec-google-news-30} & & \\
GloVe & \href{https://nlp.stanford.edu/projects/glove/}{glove.42B.300d} & & \\
\bottomrule
\addlinespace[5pt]
  \end{tabularx}
\end{table}

\end{document}